%% file: root.tex
\documentclass[letterpaper, 10 pt, conference]{ieeeconf}  

\IEEEoverridecommandlockouts                              

\usepackage{xcolor}

\usepackage{graphicx}

\usepackage{pifont}
\usepackage{gensymb}
\usepackage[hyphens]{url}
\usepackage{multirow}
\usepackage{multicol}
\usepackage{tabularx}
\usepackage{dsfont}
\usepackage{url}
\usepackage{color}
\usepackage{subcaption}
\usepackage{caption}
\usepackage{mathrsfs}
\usepackage{float}
\usepackage{booktabs}
\usepackage{dcolumn}
\usepackage{dingbat}
\let\checkmark\undefined
\usepackage{amssymb}
\usepackage{hyperref}
\usepackage{amsmath}
\usepackage[ruled,linesnumbered]{algorithm2e}

\usepackage{makecell}

\usepackage{xcolor}
\usepackage[justification=centering]{caption}

\hypersetup{
    colorlinks=true,
    linkcolor=black,
    filecolor=blue,      
    urlcolor=blue,
    pdftitle={Overleaf Example},
    pdfpagemode=FullScreen,
    }

\def\ie{\emph{i.e.}}
\def\eg{\emph{e.g.}}

\newcommand{\tiger}{\textit{TIGeR}}
\newcommand{\tigerd}{\textit{TIGeR-300K}}

\title{\LARGE \bf
TIGeR: Tool-Integrated Geometric Reasoning in Vision-Language Models for Robotics}

\author{
    Yi Han\textsuperscript{1,2,*},
    Enshen Zhou\textsuperscript{1,2,*,$\dag$}, 
    Shanyu Rong\textsuperscript{2,3}, 
    Jingkun An\textsuperscript{1}, \\
    Pengwei Wang\textsuperscript{2}, 
    Zhongyuan Wang\textsuperscript{2}, 
    Cheng Chi\textsuperscript{2,$\ddag$}, 
    Lu Sheng\textsuperscript{1,2,$\ddag$}, 
    Shanghang Zhang\textsuperscript{2,3,$\ddag$}
    \thanks{*These authors contributed equally. $\dag$Project Leader.}
\thanks{\raggedright $\ddag$Corresponding author: \texttt{chicheng@baai.ac.cn}, \texttt{lsheng@buaa.edu.cn}, \texttt{shanghang@pku.edu.cn}}
    \thanks{$^{1}$ Beihang University $^{2}$ Beijing Academy of Artificial Intelligence $^{3}$ State Key Laboratory of Multimedia Information Processing, School of Computer Science, Peking University}
}

\begin{document}

\maketitle

\begin{figure*}[ht]
\captionsetup[subfigure]{
  justification=centering,
  singlelinecheck=true}
\centering
\begin{subfigure}[ht]{0.32\linewidth}
  \centering
  \includegraphics[width=\linewidth]{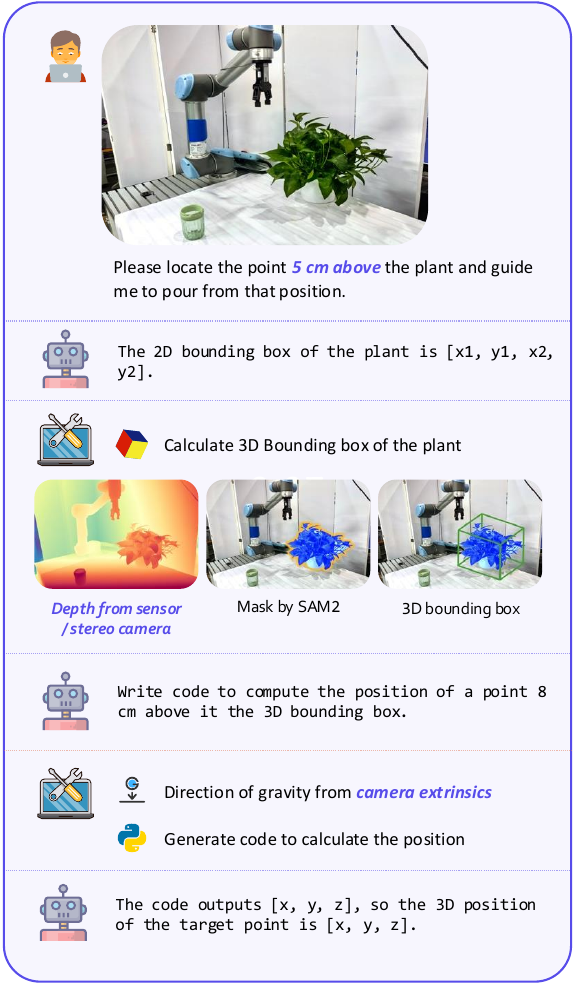}
  \caption{Accurate spatial localization}
  \label{fig:teaser:loc}
\end{subfigure}\hfill
\begin{subfigure}[ht]{0.32\linewidth}
  \centering
  \includegraphics[width=\linewidth]{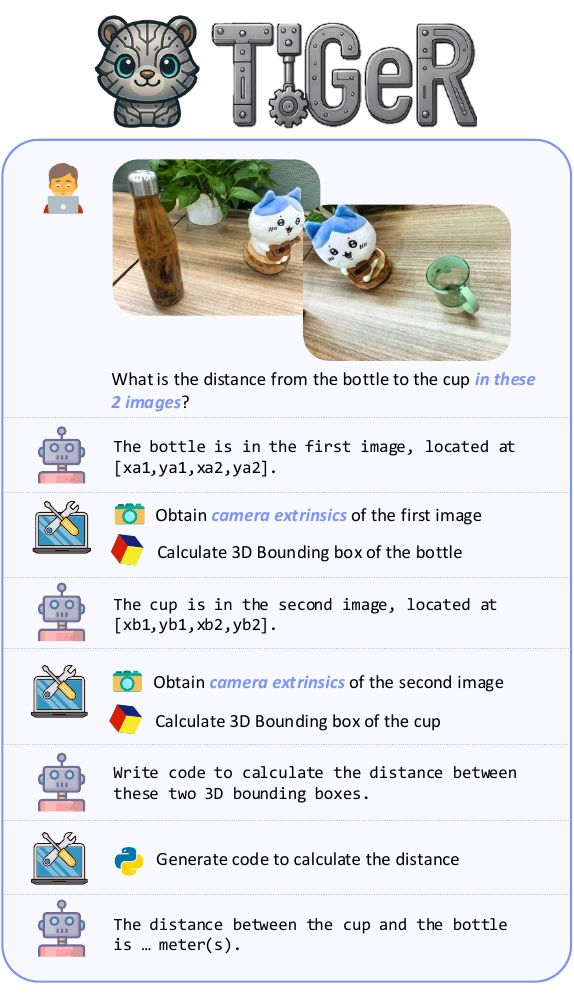}
  \caption{Unified reasoning across viewpoints}
  \label{fig:teaser:reason}
\end{subfigure}\hfill
\begin{subfigure}[ht]{0.32\linewidth}
  \centering
  \includegraphics[width=\linewidth]{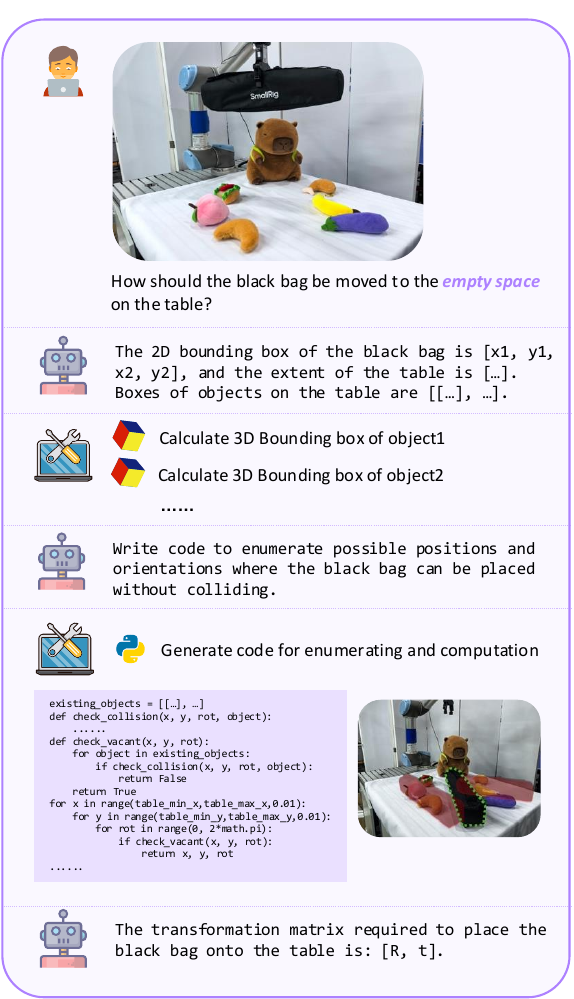}
  \caption{Executable code generation for complex computational challenges}
  \label{fig:teaser:code}
\end{subfigure}

\caption{
\tiger{} is a tool-integrated framework that enables exact geometric reasoning and computation via code generation and execution on calibrated metric inputs.
It can achieve accurate spatial localization, unified reasoning across viewpoints, and address complex embodied tasks requiring accurate numerical computation, with high interpretability and adaptability.
%
%
}
\vspace{-4mm}
\label{fig:teaser}
\end{figure*}
\thispagestyle{empty}
\pagestyle{empty}

\input{sec/0_abstract}

\input{sec/1_introduction_v2}
\input{sec/2_related_work}

\input{sec/3_method}

\input{sec/4_experiments}

\input{sec/5_conclusion}

\vspace{+1mm}
\noindent \textbf{This work was supported by the National Natural Science Foundation of China (62132001, 62476011), the Beijing Natural Science Foundation (L252060), the Fundamental Research Funds for the Central Universities, and the Xiaomi Young Talents Program/Xiaomi Foundation.}

\bibliographystyle{IEEEtran}
\bibliography{IEEEfull}

\end{document}

%% file: sec/0_abstract.tex
\begin{abstract}

Vision-Language Models (VLMs) have shown remarkable capabilities in spatial reasoning, yet they remain fundamentally limited to qualitative precision and lack the computational precision required for real-world robotics. 
Current approaches fail to leverage metric cues from depth sensors and camera calibration, instead reducing geometric problems to pattern recognition tasks that cannot deliver the centimeter-level accuracy essential for robotic manipulation. 
We present \tiger{} (Tool-Integrated Geometric Reasoning), a novel framework that transforms VLMs from perceptual estimators to geometric computers by enabling them to generate and execute precise geometric computations through external tools. 
Rather than attempting to internalize complex geometric operations within neural networks, \tiger{} empowers models to recognize geometric reasoning requirements, synthesize appropriate computational code, and invoke specialized libraries for exact calculations. 
To support this paradigm, we introduce \tigerd{}, a comprehensive tool-invocation–oriented dataset covering point transformations, pose estimation, trajectory generation, and spatial compatibility verification, complete with tool invocation sequences and intermediate computations. 
Through a two-stage training pipeline combining supervised fine-tuning (SFT) and reinforcement fine-tuning (RFT) with our proposed hierarchical reward design, \tiger{} achieves SOTA performance on geometric reasoning benchmarks while demonstrating centimeter-level precision in real-world robotic manipulation tasks. 
See the project page at \href{https://hany01rye.github.io/TIGeR/}{https://hany01rye.github.io/TIGeR/}.


\end{abstract}

%% file: sec/1_introduction_v2.tex
\section{Introduction}

Robots operating in the physical world require \textit{\textbf{geometric reasoning}}—the ability to compute metric quantities such as 3D poses, rotation matrices, and collision-free trajectories with centimeter precision. 
Yet despite recent progress, current Vision-Language Models (VLMs) remain limited to spatial reasoning~\cite{fan2025vlm, wu2025spatial, zheng2025learning}, providing only qualitative assessments of relations such as ``left of” or ``reachable”, while lacking the computational machinery for quantitative accuracy. 
This limitation prevents embodied VLMs from supporting the exact manipulations and planning that real-world robotics demands.


This limitation manifests on both the perception and output modalities. 
On the perception side, although depth sensors and stereo cameras provide rich metric information, existing models typically reduce these signals to image-like representations~\cite{cheng2024spatialrgpt, liu2025ssr, zhou2025roborefer, cai2025spatialbot}, discarding the geometric properties required for precise computation. 
Without explicit use of camera intrinsics and extrinsics, they cannot transform pixel observations into real-world coordinates, leaving them incapable of executing metric operations such as pose estimation or point cloud alignment. 
On the output side, the problem is even more pronounced. 
A few recent efforts attempt to predict 3D bounding boxes or poses~\cite{team2025gemini}, yet these results are statistical regressions trained from data rather than deterministic computations grounded in geometry. 
The majority of VLMs remain constrained to qualitative descriptions or 2D pixel predictions, which are insufficient for generating the precise poses, trajectories, and kinematic constraints that robotic manipulation and motion planning demand. 
\textit{In essence, existing approaches collapse geometric reasoning into pattern recognition, sacrificing the computational fidelity required for real-world robotics.}

To overcome this limitation, we introduce \textbf{\tiger{} (Tool-Integrated Geometric Reasoning)}, \textit{a framework that redefines the role of VLMs from perceptual estimators to geometric computers.} 
Rather than attempting to internalize geometric calculations within neural networks, \tiger{} enables models to detect when geometric reasoning is required, generate code for the appropriate operations, and invoke external tools to execute them. 
Through integration with depth sensors, camera intrinsics, and geometric libraries, VLMs can perform exact computations that significantly go beyond approximation.

To support this paradigm, we construct \textbf{\tigerd{}}, a dataset of 300K samples spanning point transformations, pose estimation, trajectory generation, and spatial compatibility checks. 
Each sample provides not only problem statements and solutions but also the complete tool invocation sequence and intermediate computations. 
Building on \tigerd{}, we develop a two-stage training pipeline: supervised fine-tuning (SFT) to instill tool usage, followed by reinforcement fine-tuning (RFT) with our proposed hierarchical reward to enhance the accuracy and task completion.

In summary, this work makes three contributions:
\begin{itemize} 

\item \textit{Concept \& Method.} We underscore the central role of \emph{geometric reasoning}—beyond qualitative spatial reasoning—for precise robotic control, and introduce \tiger{}, a tool-integrated framework that enables exact geometric computation via code generation and execution on calibrated metric inputs.


\item \textit{Dataset.} We release \tigerd{}, a large-scale, tool-invocation–oriented dataset explicitly designed for geometric reasoning via programmatic tool calls.

\item \textit{Experimental Results.} Using a two-stage SFT$\rightarrow$RFT pipeline, \tiger{} achieves state-of-the-art performance on geometric benchmarks and achieves centimeter-level accuracy in real-world robotic manipulation.

\end{itemize}

%% file: sec/2_related_work.tex
\section{Related Work}

\noindent\textbf{Tool-Integrated Reasoning.} 
Tool-Integrated Reasoning (TIR) enhances LLM and VLM reasoning ability, particularly for tasks requiring precise numerical computation, and is categorized into prompt-based, SFT-based, and RL-based methods.
Prompt-based methods~\cite{lu2025octotools, li2025search, tan2025roboos} use carefully designed prompts to invoke external tools without additional training. While easy to implement, they often exhibit low accuracy and unstable tool usage.
SFT-based methods~\cite{dong2024self, li2025you, team2025robobrain, tan2026robobrain, zhou2025robotracer, li2025robomirror} fine-tune pre-trained models on TIR-specific data, improving performance but risking memorizing tool usage patterns from training data, which may hinder generalization to unseen scenarios.
While current RL-based methods~\cite{feng2025retool, li2025torl, li2025language} promote generalization through outcome-based rewards, they overlook intermediate reasoning precision, essential for embodied tasks involving accurate numerical computation.
We thus propose novel process-based reward functions tailored for TIR and introduce a new dataset to support training.

\noindent\textbf{Spatial Understanding and Reasoning with VLMs.} 
%
Spatial understanding focuses on object-centric properties (\eg, position, orientation) and inter-object relations (\eg, distance, direction), while spatial reasoning draws higher-level inferences over such information.
Recent advances in VLMs enhance these two abilities via two paradigms: 
\textbf{(1)} Data-driven methods that fine-tune VLMs with pseudo-3D annotations~\cite{chen2024spatialvlm, cheng2024spatialrgpt}, real-world 3D datasets~\cite{song2024robospatial}, or simulated data~\cite{ray2024sat}, using either the original image encoder, an added depth encoder~\cite{cheng2024spatialrgpt, cai2025spatialbot, zhou2025roborefer} for 3D cues, or a geometry encoder~\cite{wu2025spatial} for structured scene and camera information.
While these methods may enhance spatial understanding, they struggle with embodied tasks requiring precise numerical reasoning due to VLM's hallucination and limited interpretability.
\textbf{(2)} Tool-based methods~\cite{zhou2024code} that integrate VLMs with vision foundation models to extract and reason spatial cues can achieve better performance but suffer from a limited training-free approach.
In contrast, this work introduces a sequential SFT-RFT training pipeline with novel process reward functions, to enable spatial understanding and reasoning with accurate numerical computation.

%% file: sec/3_method.tex
\section{Method}

We first give an overview of the proposed \tiger{} (Sec.~\ref{sec:overview}).
Next, we introduce the tool categorization in embodied scenarios (Sec.~\ref{sec:tool-categorization}). 
Then, we describe the construction of \tigerd{} (Sec.~\ref{sec:data-pipeline}).
Finally, we elaborate on the two-stage training pipeline (Sec.~\ref{sec:two-stage-training}), including our novel hierarchical reward design tailored for TIR.


\subsection{Overview of \tiger{}}
\label{sec:overview}

\tiger{} is a tool-integrated geometric reasoning framework that performs exact geometric computation by generating and executing code on calibrated metric inputs, enabling high-precision reasoning for downstream robotic tasks, even at centimeter-level accuracy.
Specifically, as shown in Fig~\ref{fig:teaser}(c), given a human instruction (\eg, ``\textit{How should the black bag be moved to the empty space on the table?}''), \tiger{} performs step-by-step reasoning by selecting relevant context (\eg, previous tool outputs) and invoking appropriate tools (\eg, 2D/3D detection, camera pose estimation) to obtain desired precise intermediate metric results. 
During reasoning, it can also generate code to integrate and compute over acquired information, enabling the progressive derivation of informative intermediate results and final answers.

In embodied robotics tasks demanding precise geometric reasoning, our paradigm offers three key advantages over directly using data-driven, fine-tuned, spatially-aware VLMs:
\textbf{(1) Accurate spatial localization.}
Our method enables precise prediction of positions in 3D space (\eg, 3D keypoints, trajectories) by leveraging existing 3D foundation models.
In contrast, data-driven VLMs often struggle with quantitative spatial reasoning due to training objectives (\ie, next-token prediction) and insufficient data coverage (\eg, understanding ``5cm above the plant'').
\textbf{(2) Unified reasoning across viewpoints.}
Our method is viewpoint-invariant and supports consistent numerical reasoning within a unified coordinate system, even when multi-view cameras are not jointly calibrated, enabled by recent advances in 3D geometry reconstruction models (\eg, $\pi^3$~\cite{wang2025pi}).
\textbf{(3) Interpretability and Adaptability.}
By explicitly invoking tools and exposing intermediate steps, our method offers explicit reasoning and easy integration of new tools (\eg, SOTA models). 
This enables efficient adaptation to novel tasks without costly retraining, unlike monolithic data-driven models requiring retraining to generalize to new scenarios.



\subsection{Tool Categorization in Embodied Scenarios}
\label{sec:tool-categorization}


We categorize tools into two groups: visual perception and geometric computation, reflecting a hierarchical workflow. 
Visual perception tools extract pixel- or camera-level information from sensory inputs, while geometric computation tools process this data to derive geometric properties (\eg, point transformations, pose estimation). 
The computed results are then translated into executable robot commands.
This organization enables efficient tool invocation, where the VLM determines when and how to invoke tools based on task demands, explicitly injecting precise geometric information into the reasoning process.

\textbf{Visual Perception Tools}  
extract essential sensory information to address the limitations of VLMs in metric-scale accuracy. Notably, in real-world settings where certain data (\eg, camera intrinsics and calibrated extrinsics) are readily available, we directly use these values instead of invoking additional foundation models as tools.
Key tools include:


\begin{itemize}
    \item $\mathrm{camera\_intrinsics}$: Retrieves the camera’s intrinsic parameters to support accurate 2D-to-3D projections. If unavailable, we estimate it using MoGe-2~\cite{wang2025moge}.
    \item $\mathrm{camera\_extrinsics}$: Provides extrinsic parameters to align multi-view observations within a unified 3D coordinate system. When not directly accessible, we estimate them using VGGT~\cite{wang2025vggt} or $\pi^3$~\cite{wang2025pi}, referencing the first frame as the world coordinate system.
    \item $\mathrm{depth\_sensor}$: Queries depth information from sensors for selected regions, avoiding the overhead of inputting full-depth maps to the VLM. 
    If sensor data is missing, metric depth is estimated via MoGe-2~\cite{wang2025moge}.
    \item $\mathrm{object\_segmentation}$: Generates precise segmentation masks from coarse VLM outputs (\eg, 2D bounding boxes), enabling fine-grained analysis of object shapes and boundaries.
    We use SAM2~\cite{ravi2024sam} as the tool.
\end{itemize}

\textbf{Geometric Computation Tools} perform geometric computations and transformations, bridging the gap between 2D pixel-based VLM outputs and 3D geometric requirements. 
These tools can handle embodied tasks requiring accurate numerical computation (\eg, 3D pose estimation) beyond the VLM's native capabilities.
Notably, to enhance accuracy and robustness, we introduce a code generation subroutine: the VLM specifies high-level requirements, which are fulfilled by an external code generator (Qwen3-Coder~\cite{yang2025qwen3}). The generated code, executed in a sandboxed environment, supports progressive reasoning by integrating intermediate results and producing final outputs. Key tools include:
%

\begin{itemize}
    \item $\mathrm{box\_2d\_to\_box\_3d}$: Converts a 2D bounding box (from VLM output) into a 3D bounding box by integrating 2D segmentation with depth sensor data, estimating object scale, position, and orientation in 3D space.
    \item $\mathrm{point\_3d\_to\_point\_2d}$: Projects 3D points onto the 2D image plane using camera intrinsics/extrinsics, enabling tasks such as pixel-level target location referring.
    \item $\mathrm{code\_executor}$: Invokes the code generation pipeline for arbitrary computations, such as inter-object distance calculations or pose estimations, returning numerical results to the VLM's reasoning process.
    
\end{itemize}

\begin{figure*}[t]
    \centering{\includegraphics[width=0.8\linewidth]{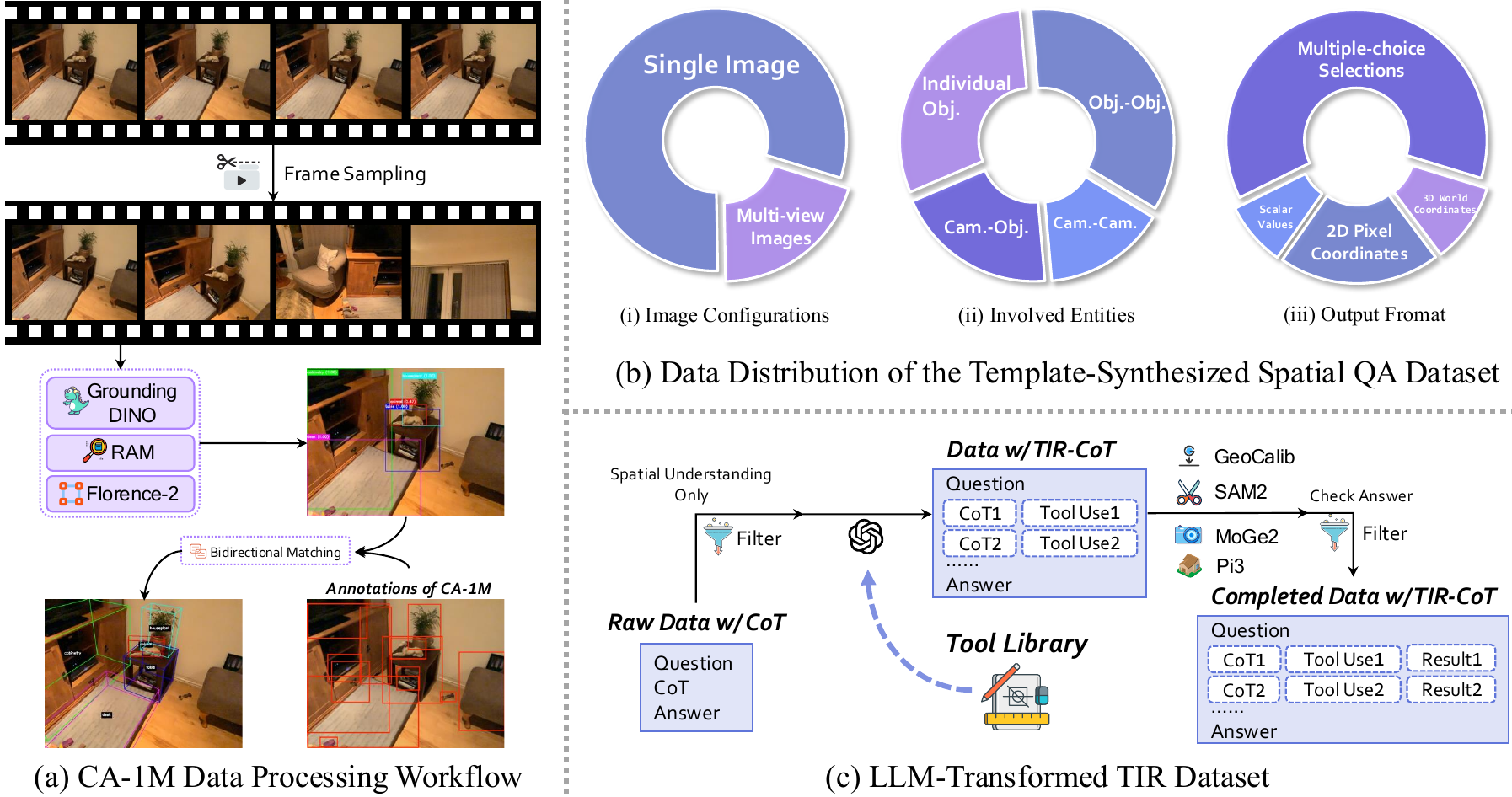}}
    \caption{\tigerd{}: A VQA dataset comprising 300K samples with Tool-Integrated Geometric Reasoning, generated via template-based synthesis and tool-augmentation by LLMs. It contains problem statements and solutions with a complete tool invocation sequence and intermediate computations tailored for our proposed hierarchical reward design.}
    \vspace{-7mm}
    \label{fig:data}
\end{figure*}

\subsection{Data Generation}
\label{sec:data-pipeline}

To support effective training of \tiger{}, we construct a large-scale dataset, \tigerd{}, comprising 300K high-quality samples that integrate tool usage with geometric reasoning in embodied scenarios.
As shown in Fig.~\ref{fig:data}, our data generation pipeline adopts a hybrid approach:
\textbf{(1)} a structured, template-based annotation method for generating precise and controllable instances targeting common spatial queries, and
\textbf{(2)} a large-model-driven chain-of-thought rewriting process to produce diverse and flexible examples that promote adaptive tool-integrated reasoning.
Each sample provides not only problem statements and solutions but also the complete tool invocation sequence and intermediate computations.
This dual strategy ensures comprehensive coverage of core geometric concepts while supporting generalization to diverse, real-world spatial and geometric reasoning tasks.


\noindent\textbf{Template-Based Data Generation.}
%
We utilize the CA-1M dataset~\cite{lazarow2025cubify} as our primary data source, which provides per-frame 2D/3D oriented bounding boxes, camera intrinsics/extrinsics, and depth maps.
To reduce redundancy, we sample every 20th frame.
However, the original dataset presents several challenges, including numerous bounding boxes with ambiguous or semantically irrelevant content, as well as a lack of semantic labels.
To address these issues, we perform data cleaning and re-annotation. 
Specifically, as shown in Fig~\ref{fig:data}(a), we leverage GroundingDINO~\cite{liu2024grounding}, RAM~\cite{zhang2024recognize}, and Florence-2~\cite{xiao2024florence} to predict semantic labels and 2D bounding boxes for previously unlabeled objects.
The predicted boxes are then matched with the original CA-1M annotations via Intersection over Union (IoU), retaining only those with high overlap to filter out noisy boxes and enrich the dataset with semantic information.
Finally, we construct 3D scene graphs—where nodes represent objects and edges encode inter-object spatial relations—by integrating cleaned 2D/3D boxes, depth maps, and camera parameters. 
These graphs serve as a foundation for downstream template-based data generation.
To address common spatial queries in robotics (\eg, localization, distance estimation, and pose computation), we design modular templates that vary systematically to generate diverse yet consistent instances, filled using annotated CA-1M frames.
%
%
%
These templates exhibit diversity (see Fig~\ref{fig:data}(b)) in the following aspects:

%

\begin{itemize}
\item \textbf{Image Configurations}: Samples comprise both single-view and multi-view images from the same scene, enabling diverse embodied task settings.
\item \textbf{Involved Entities}: Queries involve object-centric properties (\eg, size), inter-object relations (\eg, distance, spatial layout), object-camera interactions (\eg, depth), and inter-camera geometry (\eg, relative poses).
\item \textbf{Output Formats}: Answers are structured to support both robotic applications (\eg, 3D coordinates, poses) and spatial reasoning benchmarks (\eg, multiple choice, scalar values, 2D pixel locations).
\end{itemize}

\noindent All templates are instantiated using filtered and annotated CA-1M frames, ensuring access to ground-truth geometry for accurate tool simulation. 
This yields approximately 274K structured samples, forming a robust dataset for supervised learning of geometric reasoning tasks.

\noindent\textbf{Large-Model-Based CoT Rewriting.}
To improve adaptability and cover a wider range of geometric reasoning tasks, we utilize the SSR-CoT dataset~\cite{liu2025ssr}, a visual question answering (VQA) collection enriched with chain-of-thought (CoT) annotations. 
As shown in Fig.~\ref{fig:data}(c), we first apply GPT-4o to filter for spatial inference-related questions (\eg, depth estimation, object relations, and multi-view consistency), yielding a focused subset.
For each selected CoT instance, we prompt a large model to rewrite it into a TIR-enhanced version, inserting tool calls where geometric precision is needed. 
Tool responses are initially left as placeholders. 
Since SSR-CoT lacks ground-truth annotations, we invoke tools by constructing task-relevant functions using foundation models like MoGe2~\cite{wang2025moge} and GeoCalib~\cite{wang2025moge}, and populate the placeholders accordingly. 
Quality checks are performed to ensure valid tool formats and accurate final answers, resulting in 35K diverse, tool-integrated samples.

\noindent By combining template-generated and rewritten data, \tigerd{} provides the VLM with both precise supervision and compositional flexibility, enhancing its capacity to parse unstructured queries and adapt tool usage in novel scenarios.




\subsection{Two-Stage Training}
\label{sec:two-stage-training}

\begin{figure*}[t]
    \centering{\includegraphics[width=0.96\linewidth]{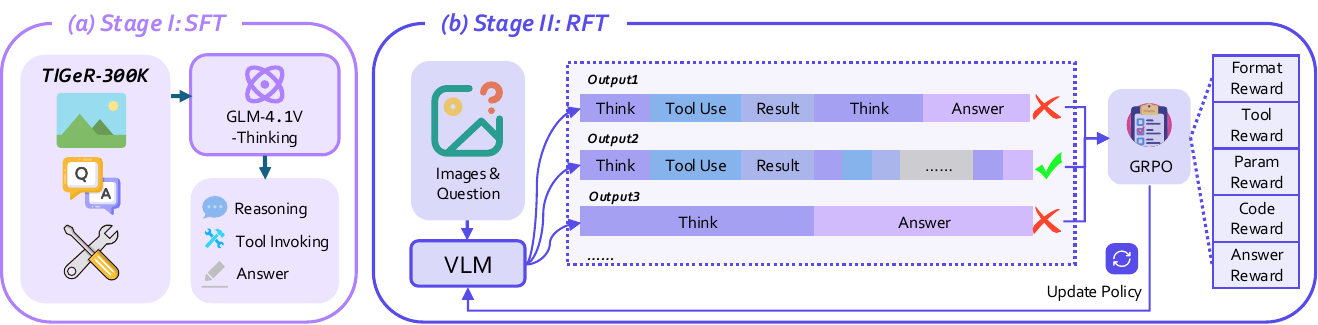}}
    \caption{\tiger{}'s two-stage training pipeline. The process begins with SFT on \tigerd{}, followed by RFT incorporating five specialized reward functions (outcome- and process-based) tailored for TIR to enhance geometric reasoning abilities.}
    \vspace{-5mm}
    \label{fig:train}
\end{figure*}


We build on GLM-4.1V-Thinking~\cite{hong2025glm} and introduce tool-integrated geometric reasoning for robotics via a two-stage training pipeline (Fig.~\ref{fig:train}): 
Supervised Fine-Tuning (SFT) followed by Reinforcement Fine-Tuning (RFT). 
Leveraging \tigerd{}, SFT imparts basic tool-use reasoning capabilities, while RFT refines them through reward signals focused on geometric computation accuracy and effective tool use.

\noindent\textbf{Cold-Start SFT.}
%
%
We first apply SFT on \tigerd{} to initialize the model's ability to generate reasoning chains with appropriate tool usage. Each sample in \tigerd{} has four elements: input image(s) $\mathcal{I}$, textual query $\mathcal{Q}$, reasoning trajectory $\mathcal{R}$ (including tool calls and code generation), and final output $\mathcal{O}$ (\eg, computed geometric results).
%
%
%
We optimize the model using a next-token prediction loss:
$$
\mathcal{L}_{\mathrm{SFT}}=-\mathbb{E}_{\tau\sim\mathcal{D}}\sum_{t=1}^T\log\pi_\theta(r_t\mid\mathbf{I},\mathbf{Q},\mathbf{r}_{<t})
$$
%
where $\mathcal{D}$ is \tigerd{}, $\pi_\theta$ denotes the parameterized VLM, $r_t$ is the token at position $t$ in the reasoning-output sequence of length $T$, and $\mathbf{r}_{<t}$ are the preceding tokens.

\noindent\textbf{RFT for Geometric Tool Use.}
We further employ RFT using GRPO~\cite{shao2024deepseekmath} to enhance the model's generalized tool-integrated geometric reasoning capabilities. 
%
%
For a given visual-language query $q$ and a batch of $N$ responses $\{y_i\}_{i=1}^N$ sampled from the current policy $\pi_\theta$, we optimize:
\begin{align*}
\mathcal{L}_{\text{GRPO}}(\theta) 
&= -\mathbb{E}_{q, \{y_i\} \sim \pi_{\theta_{\text{old}}}} \Bigg[ 
    \frac{1}{N} \sum_{i=1}^{N} 
    \min\Big( \rho_i \hat{A}_i, \nonumber \\
&\quad\quad \text{clip}(\rho_i, 1-\epsilon, 1+\epsilon)\hat{A}_i \Big) 
\Bigg] 
+ \beta D_{\text{KL}}(\pi_\theta \| \pi_{\text{SFT}})
\end{align*}
where $\rho_i = \frac{\pi_\theta(y_i|q)}{\pi_{\theta_{\text{old}}}(y_i|q)}$ is the importance ratio, $\hat{A}_i = \frac{r_i - \bar{r}}{\sigma_r}$ is the normalized advantage with reward $r_i$, mean $\bar{r}$ and standard deviation $\sigma_r$ computed over the batch, and $\beta$ controls KL regularization strength.

\noindent\textbf{Hierarchical Reward Design.} 
We design a composite reward function $r(y) = \sum_k w_k \cdot r_k(y)$ to evaluate geometric tool use from both outcome- and process-based perspectives:
\textbf{(1) Format Reward} $r_{\mathrm{format}}$ (outcome-based): Measures structural correctness of spatial tokens and tool syntax:
$r_{\mathrm{format}}(y) = \mathbb{I}[\text{valid spatial tokens}] \cdot \mathbb{I}[\text{valid tool syntax}]$
\textbf{(2) Tool Invocation Reward} $r_{\mathrm{tool}}$ (process-based): Assesses correctness of tool selection and parameter formatting:
$r_{\mathrm{tool}}(y) = \mathbb{I}[\text{tool} \in \mathcal{T}] \cdot \mathbb{I}[\text{valid parameter structure}]$
where $\mathcal{T}$ denotes the set of available geometric tools.
\textbf{(3) Parameter Content Reward} $r_{\mathrm{param}}$ (process-based): Evaluates parameter accuracy based on their type:
$$
r_{\mathrm{param}}(y) = \begin{cases}\exp(-\alpha \|p - p^*\|_2) & \text{if } p \text{ is continuous}  \\ \mathbb{I}[p = p^*] & \text{if } p \text{ is discrete} \end{cases}
$$
where $p$ are the predicted continuous (\eg, pixels, coordinates) and discrete (\eg, view indices) parameters, $p^*$ are ground-truth, and $\alpha$ is a scaling factor. 
\textbf{(4) Code Generation Reward} $r_{\mathrm{code}}$ (process-based): For code-based geometric computations:
$r_{\mathrm{code}}(y) = \lambda_{\text{exec}} \cdot \mathbb{I}[\text{code executes}] + \lambda_{\text{out}} \cdot \mathbb{I}[\text{output correct}]$
\textbf{(5) Answer Reward} $r_{\mathrm{answer}}$ (outcome-based): Assesses the final answer w.r.t. the ground truth:
$$
r_{\mathrm{answer}}(y) = \begin{cases} \exp(-\gamma \|a - a^*\|) & \text{for continuous outputs} \\\mathbb{I}[a = a^*] & \text{for discrete outputs} \end{cases} 
$$

\noindent This hierarchical reward structure provides fine-grained supervision for tool-integrated reasoning, guiding the model to learn accurate geometry computations and appropriate tool invocation, while preserving interpretable intermediate steps. 
The weights $\{w_k\}$ are tuned based on validation performance to balance structural correctness and task-level accuracy.

%% file: sec/4_experiments.tex
\section{Experiments}

Our experiments aim to address the following questions: 
(1) Can \tiger{} achieve precise spatial understanding and geometry reasoning across public benchmarks (Sec.~\ref{subsec:spatial benchmarks})?
(2) Can \tiger{} do spatially-aware manipulation tasks in simulators (Sec.~\ref{subsec:simulation}) and on real robots (Sec.~\ref{subsec:real-world evaluation}), especially requiring accurate numerical computation?
(3) Which design choices greatly enhance performance (Sec.~\ref{subsec:ablation})?

\begin{figure*}[t]
    \centering{\includegraphics[width=0.85\linewidth]{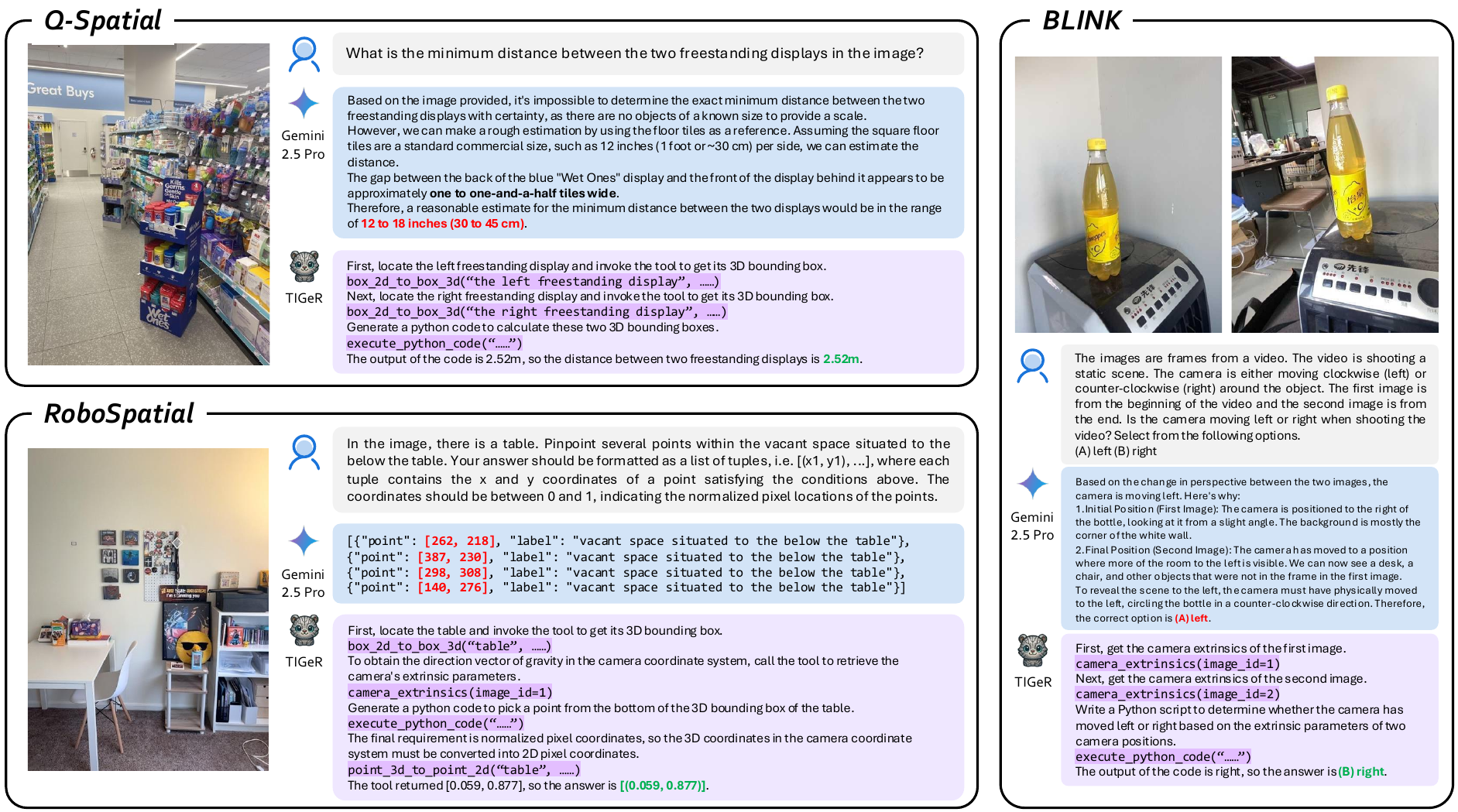}}
    \caption{Comparison between \tiger{} and Gemini 2.5-Pro on representative samples from spatial understanding benchmarks.
    }
    \vspace{-5mm}
    \label{fig:bench_samples}
\end{figure*}

\begin{table*}[t]
\centering
\caption{Performance comparison on spatial understanding and geometry reasoning benchmarks across different models. M.V. indicates Multi-view parts; Conf., Comp., Cont. denote the configuration, compatibility, context part of RoboSpatial. Top-1 \& Top-2 accuracies are represented using \textbf{bold text}, and \underline{underlines}. For Q-Spatial++, the metric is $\delta_{\le 2}$, where the answer is considered correct if it falls within $[0.5\times, 2\times]$ of the ground truth.}
\vspace{+0mm}
\label{tab:performance_comparison}
\scriptsize
\begin{tabular}{lcccccccccccc}
\toprule
\multirow{2}{*}{Models} &
\multicolumn{3}{c}{CV-Bench~\cite{tong2024cambrian}} &
\multicolumn{3}{c}{BLINK~\cite{fu2024blink}} &
\multicolumn{3}{c}{RoboSpatial~\cite{song2025robospatial}} &
\multirow{2}{*}{EmbSpatial~\cite{du2024embspatial}} &
\multirow{2}{*}{Q-Spatial++~\cite{liao2024reasoning}} \\
\cmidrule(lr){2-4}\cmidrule(lr){5-7}\cmidrule(lr){8-10}
 & 2D-Rel & 3D-Depth & 3D-Dist  
 & Depth & Spatial & M.V. 
 & Conf. & Comp. & Cont.   
 \\ \midrule
GPT-4o~\cite{achiam2023gpt} & 84.62 & 86.50 & 83.33 & 78.23 & 82.52 & 51.88 & 77.20 & 61.90 & 25.30 & 63.38 & 20.79 \\
Gemini-2.5-Pro~\cite{team2023gemini} & \underline{93.54} & 91.00 & \underline{90.67} & 87.90 & \textbf{91.61} & 35.34 & 77.24 & \underline{68.57} & \underline{40.20} & \underline{76.67} & 55.46 \\
\midrule 
NVILA-8B~\cite{liu2025nvila} & 91.54 & \underline{91.83} & \underline{90.67} & 76.61 & 76.92 & \underline{55.64} & 59.35 & 23.81 & 5.51 & 67.72 & 50.5 \\
Qwen-2.5-VL-7B~\cite{bai2025qwen2} & 82.15 & 60.17 & 69.00 & 60.98 & 64.34 & 48.12 & 49.59 & 66.67 & 13.61 & 40.20 & 49.50 \\
GLM-4.1V-Thinking~\cite{hong2025glm} & 93.08 & 92.00 & 88.33 & \underline{90.32} & 81.82 & 47.37 & \underline{81.30} & 43.80 & 8.24 & 76.62 & \underline{67.33} \\
\midrule
SpatialBot-3B~\cite{cai2025spatialbot} & 69.38 & 77.33 & 60.83 & 67.74 & 67.83 & 44.36 & 72.36 & 22.86 & 4.88 & 50.66 & 11.88 \\
SpaceLLaVA-13B~\cite{foutter2024space} & 63.69 & 66.83 & 70.17 & 62.90 & 72.73 & 43.61 & 61.00 & 38.10 & 16.00 & 49.40 & 37.62 \\
RoboPoint-13B~\cite{yuan2024robopoint} & 75.85 & 77.83 & 44.5 & 61.29 & 60.84 & 44.36 & 69.9 & 21.90 & \textbf{41.30} & 49.31 & 20.79 \\
\midrule
\tiger{} & \textbf{93.85} & \textbf{96.33} & \textbf{95.17} & \textbf{91.94} & \underline{86.01} & \textbf{60.15} & \textbf{82.11} & \textbf{82.86} & 32.79 & \textbf{80.82} & \textbf{70.30} \\
\bottomrule
\vspace{-9mm}
\end{tabular}
\end{table*}

\subsection{Spatial and Geometry Reasoning Benchmarks}
\label{subsec:spatial benchmarks}


We evaluate \tiger{} on qualitative spatial understanding benchmarks, including CV-Bench~\cite{tong2024cambrian}, BLINK~\cite{fu2024blink}, RoboSpatial~\cite{song2025robospatial}, and EmbSpatial~\cite{du2024embspatial}, as well as the quantitative geometric reasoning benchmark Q-Spatial++~\cite{liao2024reasoning}. 
Since these benchmarks lack ground-truth geometric annotations (\eg, camera intrinsics, extrinsics, and depth), we leverage visual foundation models (see Sec.~\ref{sec:tool-categorization}) to extract such information and inject approximate geometric priors into Tool-Integrated Reasoning at inference.
As shown in Tab.~\ref{tab:performance_comparison}, \tiger{} achieves a zero-shot state-of-the-art accuracy of 79.30\% across these benchmarks, even outperforming Gemini 2.5-Pro by 5.83\%.
Fig.~\ref{fig:bench_samples} further illustrates representative examples where \tiger{} successfully leverages tool-integrated reasoning to solve complex spatial and geometric problems.
We find that our framework can not only enhance its spatial understanding, enabled by tool-use during reasoning, but also improve its geometry computation accuracy, powered by code generation and numerical calculation.

%



\subsection{Simulation Evaluation for Robotics}
\label{subsec:simulation}

We evaluate our model on the  Open6DOR V2 Level 0 and Level 1 of position track, comparing against VLA-based baselines (pretrained Octo, LIBERO-finetuned OpenVLA) and SoFar, which integrates Florence-2, SAM, GPT-4o, and GSNet.
This benchmark requires models to understand different spatial concepts (\eg, ``left'', ``front'', ``between'') and manipulate objects.
%
We find that, by predicting a single 3D target point compared to end-to-end VLA models or 2D bounding box-based methods, \tiger{} mitigates 2D detection ambiguity under occlusion, yielding an 11.3\% absolute improvement in success rate (Tab.~\ref{tab:sim}) against these baselines.



\begin{figure*}[t]
    \centering{\includegraphics[width=0.90\linewidth]{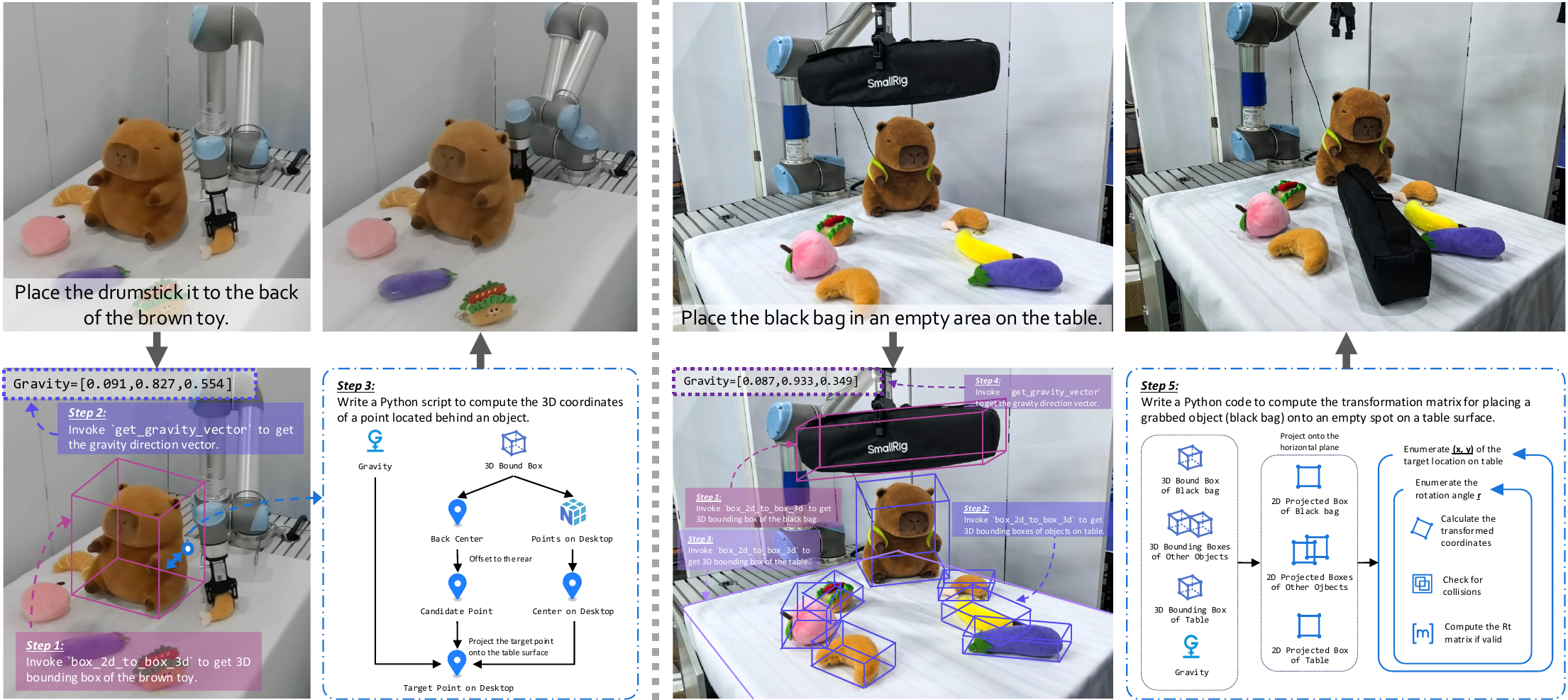}}
    \caption{Illustration of \tiger{}’s tool-integrated reasoning in real-robot tasks..
    ``\textit{Pick up the drumstick and place it to the back of the brown toy}'' requires precise 3D spatial reasoning to localize the target position despite occlusion of the rear region.
    Similarly, ``\textit{Place the black bag in an empty area on the table}'' demands complex geometric computation to identify feasible free space in cluttered scenes.
    We visualize the robot’s execution through sequential video frames, along with \tiger{}’s step-by-step tool calls and their outputs, overlaid on the corresponding images.
    }
    \vspace{-7mm}
    \label{fig:exp_sample}
\end{figure*}

\begin{table}[t]
\centering
\caption{Simulation results on Open6DOR V2 position track (Level 0 and Level 1). We report the success rate (\%).}
\vspace{+2mm} 
\begin{tabular}{l|ccc|c}
\toprule
Part & Octo & OpenVLA & SoFar & \tiger{}
\\ \midrule
Level 0 & 51.2 & 51.6 & 75.3 & \textbf{84.8} \\
Level 1 & 12.7 & 13.1 & 65.6 & \textbf{81.0} \\
Avg.    & 39.8 & 40.2 & 72.4 & \textbf{83.7} \\
\bottomrule
\end{tabular}
\label{tab:sim}
\vspace{-8mm}
\end{table}

\begin{figure}[t]
    \centering{\includegraphics[width=0.99\linewidth]{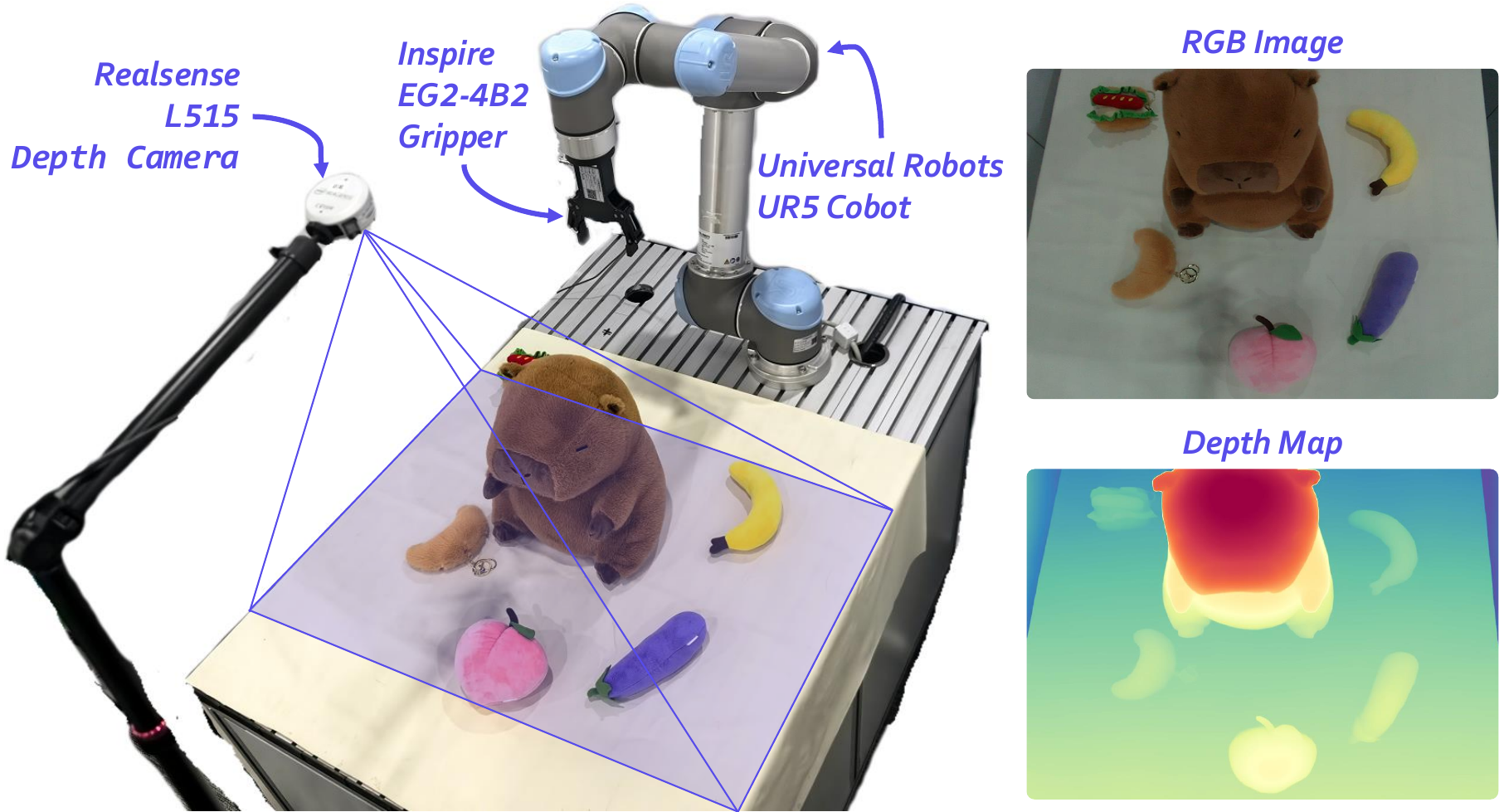}}
    \caption{Real-world settings, including the UR5 robotic arm, L515 RGB-D camera, Inspire EG2-4B2 gripper. We show the captured RGB image with the depth map.}
    \vspace{-8mm}
    \label{fig:exp_settings}
\end{figure}

\begin{table}[t]
\centering
\scriptsize
\setlength{\tabcolsep}{1pt}
\caption{Real-world evaluation of manipulation tasks requiring precise geometry reasoning. We report success rates (\%). For the task ``Pick up the hotdog and place it 0.1 m to the right of the peach'', success is defined as the actual placement distance being between 0.05 m and 0.15 m.
}
\vspace{+2mm}
\begin{tabular}{l|ccc}
\toprule                          
\multirow{2}{*}{Manipulation tasks of 3D spatial positioning} & \multicolumn{3}{c}{Success Rate(\%) $\uparrow$} \\
                      & OpenVLA & RoboPoint & \tiger{} \\
                      
\midrule
Pick up the croissant and place it to the \textbf{left} & \multirow{2}{*}{0.00} & \multirow{2}{*}{65.00} & \multirow{2}{*}{\textbf{75.00}} \\
 of the drumstick. & & &\\
 
\midrule
Pick up the hotdog and place it \textbf{0.1 m} to the & \multirow{2}{*}{0.00} & \multirow{2}{*}{10.00} & \multirow{2}{*}{\textbf{55.00}} \\
right of the peach. & & & \\

\midrule
Pick up the drumstick and place it to the \textbf{back}  & \multirow{2}{*}{0.00} & \multirow{2}{*}{0.00} & \multirow{2}{*}{\textbf{60.00}} \\
of the brown toy (\textit{facing occlusion of rear region}). & & & \\

\midrule
Pick up the bottle and move it \textbf{above} the plant& {0.00} & {0.00} & {\textbf{70.00}} \\

\bottomrule[1pt]
\end{tabular}
\label{tab:realworldexp}
\vspace{-8mm}
\end{table}

\begin{table}[t]
\caption{
Ablation studies on data recipes and reward design. We evaluate the impact of using template-based data (T.B.) \textit{vs.} LLM-rewritten data (Rewritten.), and the individual effects of five reward functions.
Tool. and Param. denote the tool invocation reward and parameter content reward.
%
%
The last column reports the average accuracy (\%) across geometric reasoning benchmarks described in Sec.~\ref{subsec:spatial benchmarks}.
}
\footnotesize
\centering
\setlength{\tabcolsep}{1.5pt}
\begin{tabular}{cc|ccccc|c}
\toprule                          
\multicolumn{2}{c}{Data Recipe} & \multicolumn{5}{c}{Reward} & Avg. \\
\cmidrule(lr){1-2} \cmidrule(lr){3-7}
T.B. & Rewritten. & Format & Tool. & Param. & Code & Answer & Acc.(\%) \\
\midrule
\ding{55} & \checkmark & \checkmark  & \checkmark  & \checkmark & \checkmark & \checkmark & 28.92 \\
\checkmark & \ding{55} & \checkmark  & \checkmark  & \checkmark & \checkmark & \checkmark & 74.46 \\
\checkmark & \checkmark & \ding{55}  & \checkmark  & \checkmark & \checkmark & \checkmark & 78.81 \\
\checkmark & \checkmark & \checkmark  & \ding{55}  & \checkmark & \checkmark & \checkmark & 76.12 \\
\checkmark & \checkmark & \checkmark  & \checkmark  & \ding{55} & \checkmark & \checkmark & 75.96 \\
\checkmark & \checkmark & \checkmark  & \checkmark  & \checkmark & \ding{55} & \checkmark & 79.11 \\
\checkmark & \checkmark & \checkmark  & \checkmark  & \checkmark & \checkmark & \ding{55} & 78.39 \\
\checkmark & \checkmark & \checkmark  & \checkmark  & \checkmark & \checkmark & \checkmark & \textbf{79.30} \\
\bottomrule
\end{tabular}
\label{tab: ablation}
\vspace{-8mm}
\end{table}

\subsection{Real-world Evaluation for Robotics}
\label{subsec:real-world evaluation}

Our real-world setting is shown in Fig.~\ref{fig:exp_settings}.
We deploy the model on a UR5 robotic arm equipped with an Intel RealSense L515 RGB-D camera.
We evaluate \tiger{} on a suite of spatially-aware tasks that require geometry reasoning and high-precision 3D localization, which are infeasible for standard 2D pixel-based VLMs due to their inability to infer depth, camera extrinsics, or occluded/hovering positions. 
Specifically, we use AnyGrasp~\cite{fang2023anygrasp} to predict a valid grasp pose and further estimate the target’s 3D location in the camera frame to guide the placement.



We evaluate our method on four challenging real-world, spatially-aware manipulation tasks, with results summarized in Tab.~\ref{tab:realworldexp}. These tasks require fine-grained spatial reasoning, such as metric-precision placement (\eg, ``0.1m to the right'') and handling partial occlusions (\eg, the ``back'' of an object being invisible to the camera).
Our method, \tiger{}, consistently outperforms all baselines. It achieves a 55\% success rate in metric-precision placement, demonstrating its ability to accurately map spatial constraints to 3D coordinates.
For spatial relations such as ``back'' and ``above'', \tiger{} achieves 60–70\% success rates, significantly higher than baselines lacking depth-aware reasoning.
Moreover, as shown in Fig.~\ref{fig:exp_sample}, \tiger{} is the only method capable of resolving partial occlusions through precise geometric computation, highlighting the effectiveness of our explicit 3D coordinate representation for real-world manipulation in complex environments.

\subsection{Ablation Study}
\label{subsec:ablation}

We conduct comprehensive ablation studies to analyze the contribution of each component in our framework, as shown in Tab.~\ref{tab: ablation}. 
For the data recipe, we compare template-based synthetic data with LLM-rewritten data, finding that combining both achieves the best performance (79.30\%), while using only rewritten data significantly degrades accuracy to 28.92\%. 
This validates the importance of structured template data for establishing robust geometric reasoning foundations. 
For the reward design in RFT, we evaluate five reward components and find that removing any single reward leads to performance degradation. 
These results demonstrate that both diverse training data and comprehensive reward signals are essential for enabling reliable tool-integrated geometric reasoning in VLMs.

%% file: sec/5_conclusion.tex
\section{conclusion}

In this work, we present \tiger{}, a novel tool-integrated framework that achieves precise geometric reasoning beyond spatial reasoning.
In detail, we propose a two-stage SFT-RFT training pipeline with our proposed hierarchical reward design, enabling exact geometric computation via code generation and execution on calibrated metric inputs during the reasoning process.
We also introduce \tigerd{}, a large-scale dataset explicitly designed for geometric reasoning via programmatic tool calls, with a complete tool invocation sequence and intermediate computations.
Extensive experiments demonstrate the superiority of the proposed approach and highlight its potential for a broad range of robotic applications requiring precise numerical computation.
